\documentclass[11pt,a4paper]{article}
\usepackage[hyperref]{AACL-IJCNLP2020}
\usepackage{times}
\usepackage{latexsym}

\usepackage[whole]{bxcjkjatype}

\usepackage{microtype}

\aclfinalcopy 

\usepackage{booktabs}
\usepackage{graphicx}
\usepackage{amsmath}
\usepackage{amsfonts}
\usepackage{amssymb}
\usepackage[font=small]{caption}
\usepackage{bm}
\usepackage{comment}
\usepackage[subrefformat=parens]{subcaption}
\usepackage{CJKutf8}
\usepackage{threeparttable}

\usepackage{here}

\title{Text Classification through Glyph-aware Disentangled Character Embedding and Semantic Sub-character Augmentation}

\author{
    Takumi Aoki \quad Shunsuke Kitada \quad Hitoshi Iyatomi \\
    Department of Applied Informatics, Graduate School of Science and Engineering \\
    Hosei University, Tokyo, Japan \\
    {\tt \{takumi.aoki.4g, shunsuke.kitada.8y\}@stu.hosei.ac.jp} \\
    {\tt iyatomi@hosei.ac.jp}
}
\date{}

\begin{document}
\maketitle
\begin{abstract}
    We propose a new character-based text classification framework for non-alphabetic languages, such as Chinese and Japanese.
Our framework consists of a variational character encoder (VCE) and character-level text classifier.
The VCE is composed of a $\beta$-variational auto-encoder ($\beta$-VAE) that learns the proposed glyph-aware disentangled character embedding (GDCE).
Since our GDCE provides zero-mean unit-variance character embeddings that are dimensionally independent, it is applicable for our interpretable data augmentation, namely, semantic sub-character augmentation (SSA). 
In this paper, we evaluated our framework using Japanese text classification tasks at the document- and sentence-level.
We confirmed that our GDCE and SSA not only provided embedding interpretability but also improved the classification performance.
Our proposal achieved a competitive result to the state-of-the-art model while also providing model interpretability.
\end{abstract}

\section{Introduction}
    Some Asian languages (e.g., Chinese and Japanese) use \textit{glyphs} to give visual meaning to characters.
For example, the following Japanese characters have a common form of \begin{CJK*}{UTF8}{ipxm}``辶,''\end{CJK*} which is a sub-character meaning of the related word \textit{road}: ``迫'' (approach: come near the destination by \textit{road}) and ``追'' (follow: track the \textit{road}).
In consideration of these characteristics of the language, several glyph-aware natural language processing (NLP) models have been proposed~\cite{shimada2016document, liu2017learning, kitada2018end, sun2019vcwe}.
These deep-learning-based models train input text as a sequence of character images and learn character embeddings from the images.

In general, the interpretability of the NLP model is important in terms of its reliability, as well as providing the required performance for the task.
If imaged-based models can learn these sub-characters in a way that is interpretable, it helps greatly in improving the overall interpretability of the models.

In terms of improving the interpretability of models, disentangled representation learning method has received a great deal of attention in recent years, such as InfoGAN~\cite{chen2016infogan} and $\beta$-variational auto-encoder ($\beta$-VAE)~\cite{higgins2017beta}.
This learning method transforms the input data into low-dimensional representations that are independent of each other while still retaining the important content.
Although it has been actively discussed in the field of computer vision, there are few applications in the field of NLP.

In terms of ensuring model robustness, data augmentation is necessary and essential in machine learning today.
With regard to this desirable feature, glyph-aware embedding (i.e., image-based character embedding) allows data augmentation without contextual consideration, such as word dropout~\cite{iyyer2015deep} and wildcard training~\cite{shimada2016document}.
Simple data augmentation based on dropout does not consider the features of the input space.
If the NLP method based on glyph-aware embedding is highly interpretive, such as a disentangled representation, an effective data augmentation method can be achieved.
This improves not only the robustness of the model but also its interpretability.

In this paper, we propose a general-purpose text classification framework that gives interpretability to data augmentation for image-based glyph-aware character embedding, which has the various advantages mentioned above.
The framework consists of two novel methods:  (1) glyph-aware disentangled character embedding (GDCE) and (2) semantic sub-character augmentation (SSA).
Each method has the following simple but effective features:
\begin{itemize}
    \item The GDCE is obtained from the variational character encoder (VCE), which is the encoder part of the $\beta$-VAE. The VCE takes advantage of the  $\beta$-VAE to create a low-dimensional representation of the characters, where each dimension follows an independent normal distribution. Therefore, the GDCE provides a disentangled character embedding in which each of the dimensions corresponds to the structure of the sub-character.
    \item The SSA alters only one dimension of the GDCE, which corresponds with altering some part of the shape of the original character, and can present how the character has changed. In other words, these combinations are equivalent to replacing the sub-character of a character with another readable sub-character. 
\end{itemize}
Our framework improves the interpretability of character embedding by the GDCE, and the SSA provides interpretable data augmentation suitable for the GDCE.
We verified the text classification ability of our proposed framework using Japanese text classification tasks. 
\footnote{
The code required to reproduce the experiments is available on GitHub. \url{https://github.com/IyatomiLab/GDCE-SSA}}

\section{Related work}
    \subsection{Glyph-aware Natural Language Processing}
Embedding methods based on character images have been proposed with some excellent success~\cite{chen2015joint, sun2016inside, yu2017joint, sun2019vcwe, dai2017glyph, shimada2016document, liu2017learning, kitada2018end, ke2017radical, aldon2016neural}.
These methods are also called glyph-aware embedding as they generate embeddings that take into account the shape of the characters or sub-characters.
These image-based methods mainly use convolutional neural networks (CNNs) or convolutional auto-encoders (CAEs)~\cite{masci2011stacked} for character-embedding learning, and they perform well because of the following advantages: (1) they operate without the cumbersome word segmentation required by some Asian languages, and (2) they can apply additional image-based data augmentation.

\subsection{Data Augmentation for Natural Language Processing}
For NLP tasks, it is challenging to apply data augmentation methods because of the need to consider the context of the text~\cite{sennrich2016improving, jia2016data, silfverberg2017data, edunov2018understanding}.
Several data augmentation methods that do not require text analysis have been proposed for word embedding~\cite{iyyer2015deep, zhang2016learning} and character embedding~\cite{shimada2016document}.
In particular, \citet{shimada2016document} achieved significant performance improvements by applying dropout~\cite{hinton2012improving}-based data augmentation to a type of character embedding called wildcard training (WT).
However, these methods have little interpretability of what the data augmentation means in the input text, partly due to the lack of interpretability of the embedding itself. 
Our proposed SSA is improved WT, and it replaces the sub-character of a character with another readable sub-character.

\subsection{Learning Interpretable Character Embeddings}
For learning a latent representation that can be interpreted, InfoGAN~\cite{chen2016infogan} and $\beta$-VAE~\cite{higgins2017beta} are well known.
Unlike InfoGAN, $\beta$-VAE is stable while training, requires less assumptions about the data, and relies on only a single hyperparameter $\beta$.
Because of these advantages, several improved models based on $\beta$-VAE have been proposed (e.g., Factor-VAE~\cite{kim2018disentangling}, HFVAE~\cite{esmaeili2019structured}).
Therefore, in this paper, we use $\beta$-VAE as a VCE to learn interpretable character embeddings.

\section{Methodology}
\begin{figure}[t]
    \centering
    \includegraphics[width=\linewidth]{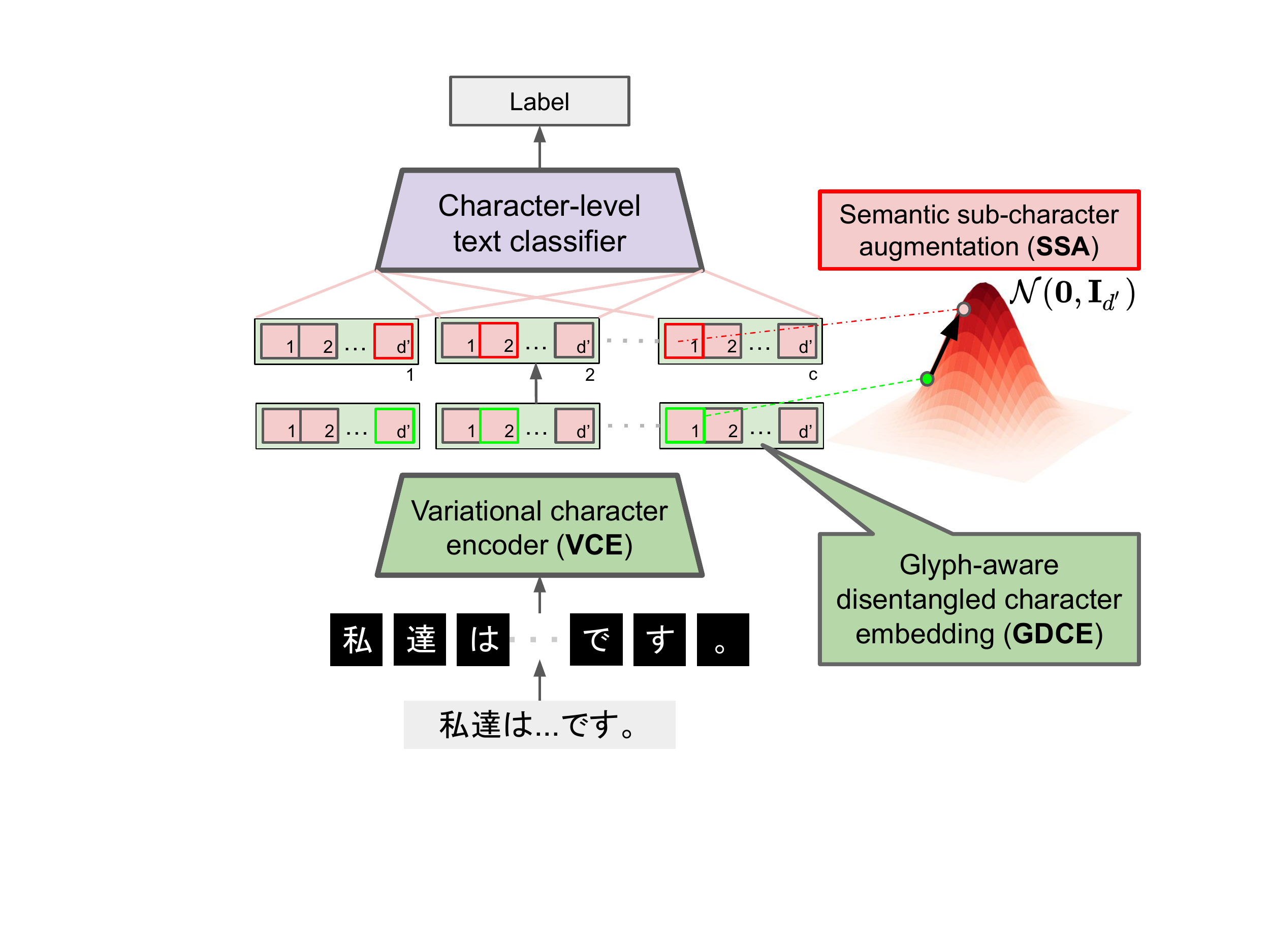}
    \caption{
        Overview of our text classification framework.
        Each character in the target text is transformed to an image and forwarded as a glyph feature to the subsequent VCE.
        The VCE is composed of a $\beta$-VAE, and it learns the proposed GDCE.
        Owing to the attractive properties of the GDCE, character-level text classifier can take advantage of the interpretable and highly effective data augmentation method, SSA.
    }
    \label{fig:method}
\end{figure}

    In this paper, we propose a new character-based text classification framework that includes a new character embedding method, consisting of glyph-aware disentangled character embedding~(GDCE) and semantic sub-character augmentation (SSA).
Figure~\ref{fig:method} shows an overview of the proposed text classification framework.

\subsection{Glyph-aware Disentangled Character Embedding (GDCE)}
We obtain the GDCE using the VCE based on the $\beta$-VAE.
Since the GDCE provides dimensionally independent features, we expect to solve the problem of the poorly interpretable character embedding obtained by the CAE.

$\beta$-VAE is a generative model that estimates the data distribution $p(\bm{x})$, where $\bm{x} \in \mathbb{R}^{d}$ is a $d$-dimensional input.
Let $\bm{z} \in \mathbb{R}^{d'}$ be a $d'$-dimensional latent variable, which is derived from the GDCE in this paper; $p(\bm{z})$ is a normal distribution, which is the prior distribution of the latent variables, $q(\bm{z}|\bm{x})$ is the posterior distribution, and $p(\bm{x}|\bm{z})$ is a generative model.
We optimize the following function:
\begin{equation}
\begin{split}
    \mathcal{L}_{\beta{\textrm{-VAE}}} ={}& \mathbb{E}_{q(\bm{z}|\bm{x})}[\log{p(\bm{x}|\bm{z})}] \\
    &- \beta D_{\mathrm{KL}}[q(\bm{z}|\bm{x})||p(\bm{z})],
    \label{eq:loss_vae}
\end{split}
\end{equation}
where $\beta$ is a balancing coefficient for the second term.
The first term represents the reconstruction error of the character image.
The second term represents the regularization of the latent variables that are learned so as to follow the prior distribution by the KL divergence $D_{\mathrm{KL}}[\cdot||\cdot]$.
If the coefficient $\beta$ increases, it is possible to obtain a representation of the features where each dimension is independent~\cite{higgins2017beta}.

However, the latent variables themselves are a probability distribution and cannot be backpropagated to the encoder.
Hence, the reparameterization trick~\cite{kingma2013auto} of the approximation method is used.
We let $\bm{\alpha}$ be a sampled random variable from $\mathcal{N}(\bm{0},\bm{I}_{d'})$ and calculate the latent variables as follows:
\begin{equation}
    \begin{array}{ll}
    \bm{z} = \mu(\bm{x}) + \bm{\alpha} \odot \sigma(\bm{x}), & \bm{\alpha} \sim \mathcal{N}(\bm{0},\bm{I}_{d'}),
    \end{array}
\end{equation}
where $\odot$ is an element-wise product, $\mu$ is the mean of the distribution, and $\sigma$ is the variance of the distribution.
Here, $\mu(\bm{x})$ and $\sigma(\bm{x}) $ are $d'$-dimensional vectors obtained from the $\beta$-VAE.

\subsection{Character-level Text Classification with Semantic Sub-character Augmentation (SSA)}
The sequence of $c$ embedded characters $C = \{\bm{z}^{(1)}, \bm{z}^{(2)}, \cdots, \bm{z}^{(c)}\}$ from the GDCE, where $\bm{z}^{(t)}$ is the $t$-th character embedding, in the VCE are provided to the following character-level text classifier.
The parameters of the classifier are optimized in the back-propagation using the cross-entropy error.

In this paper, we propose SSA as a data augmentation method.
Taking advantage of the preferred features of the embedding created by the GDCE, we expect that the sub-character of a character will be replaced by another readable sub-character, using the SSA.

Let $\gamma$ be the perturbation range, and the formula of the SSA for the $i$-th dimension $\bm{z}^{(t)}_i$ of the character embedding $\bm{z}^{(t)}$ is defined as follows:
\begin{equation}
    \begin{array}{ll}
    \bm{z'}^{(t)}_{i} = \bm{z}^{(t)}_{i} + u, & u \sim \mathcal{U}(-\gamma, \gamma),
    \label{eq:iwt}
    \end{array}
\end{equation}
where $u \sim \mathcal{U}(a, b)$ indicates that the random variable $u$ has a uniform distribution with the minimum $a$ and the maximum $b$.
Since each dimension of the GDCE follows $\mathcal{N}(\bm{0}, \bm{I}_{d'})$, the character embedding converted in Eq.~\ref{eq:iwt} falls within the range of trained character-embedding values.
    
\section{Experiment Settings}
    \subsection{Evaluation Datasets}
We evaluated our framework with the following datasets: newspaper and livedoor.
These datasets were split into two parts: 80\% for training and 20\% for evaluation.
Because these datasets contain new words and/or meanings related to current affairs, accurate word segmentation through morphological analysis has been a challenge in conventional word-level processing for Japanese.
Therefore, we can avoid such difficulties by using character-level input instead of word-level input\footnote{It is generally known that a character-level model performs better than a word-level model in Chinese and Japanese~\cite{zhang2017encoding}.}.

\paragraph{Newspaper.} The newspaper dataset used in \citet{shimada2016document} contains 5,610 Japanese major web newspaper articles (Asahi, Mainichi, Sankei, and Yomiuri) in the categories of politics, the economy, and international news, for a total of 22,440 articles.

\paragraph{Livedoor.} The livedoor dataset is commonly used to evaluate models for Japanese.\footnote{\url{https://www.rondhuit.com/download.html\#ldcc}}
The dataset contains, for example, 870 and 900 Japanese sentences in the categories of movie-enter and sports-watch, respectively.
In all the nine categories, it contains a total of 7,367 articles.

\subsection{Model Architectures}
We trained the VCE based on $\beta$-VAE and character-level CNN (CLCNN)~\cite{zhang2015character} as text classifier independently.
The hyperparameters of these models were adjusted with a validation set split from the training set, and the predicted results of the evaluation set were reported.

\begin{table}[t]
    \centering
    \begin{minipage}{\linewidth}
        \centering
{\footnotesize
\begin{tabular}{@{}cc@{}}
\toprule
Layer & Encoder                                          \\ \cmidrule(r){1-1} \cmidrule(lr){2-2}
1     & Conv2d ($k=(4, 4)$, $o=32$, $s=2$) $\rightarrow$ ReLU     \\
2     & Conv2d ($k=(4, 4)$, $o=32$, $s=2$) $\rightarrow$ ReLU     \\
3     & Conv2d ($k=(4, 4)$, $o=64$, $s=2$) $\rightarrow$ ReLU     \\
4     & Conv2d ($k=(4, 4)$, $o=64$, $s=2$) $\rightarrow$ ReLU     \\
5     & Linear($o=256$) $\rightarrow$ ReLU                 \\
6     & Linear($o=2\times10$)                          \\ \bottomrule
\end{tabular}%
}
\label{tab:vae-encoder}
    \end{minipage}
    \\
    \vspace{2.5mm} 
    
    \begin{minipage}{\linewidth}
        \centering
{\footnotesize
\begin{tabular}{@{}cc@{}}
\toprule
Layer & Decoder                                        \\ \cmidrule(r){1-1} \cmidrule(lr){2-2}
1     & Linear ($o=256$) $\rightarrow$ ReLU               \\
2     & Linear ($o=1024$) $\rightarrow$ ReLU              \\
3     & Deconv2d ($k=(4, 4)$, $o=64$, $s=2$) $\rightarrow$ ReLU    \\
4     & Deconv2d ($k=(4, 4)$, $o=32$, $s=2$) $\rightarrow$ ReLU    \\
5     & Deconv2d ($k=(4, 4)$, $o=32$, $s=2$) $\rightarrow$ ReLU    \\
6     & Deconv2d ($k=(4, 4)$, $o=1$, $s=2$) $\rightarrow$ Sigmoid \\ \bottomrule
\end{tabular}%
}
\label{tab:vae-decoder}
    \end{minipage}\\
    
    \caption{
        Architecture of $\beta$-VAE.
        Kernel size $k$, output size $o$, and stride size $s$ was set to the above table.
    }
    \label{tab:arch_vae}
\end{table}

\paragraph{$\beta$-variational auto-encoder ($\beta$-VAE).}
Table~\ref{tab:arch_vae} shows the architecture of $\beta$-VAE.
Generally, training of $\beta$-VAE is unstable, and requires adjustment of hyperparameters.
In this paper, we carefully tuned hyperparameters based on \citet{locatello2019challenging}.
Adam~\cite{kingma2014adam} was used to maximize $\mathcal{L}_{\beta{\textrm{-VAE}}}$, as shown in Eq.~\ref{eq:loss_vae}.
We set train batch size to 64 and the learning rate to 1e-4.

To obtain the GDCE, we trained the VCE with 6,631 common Japanese characters, including Japanese Hiragana, Katakana, and Kanji\footnote{From the Japanese Industrial Standards; first and second levels.}, as well as English alphabets and symbols.
These characters were converted to $d = 64 \times 64$ grayscale character images and used as input $\bm{x}$ to the VCE.
We set $\beta = 8$ and $d' = 10$ for all tasks, $\gamma = 1.5$ for the newspaper, and $\gamma = 2.0$ for the livedoor.

\begin{table}[t]
\centering
{\footnotesize
\begin{tabular}{@{}cc@{}}
\toprule
Layer & CLCNN                                  \\ \cmidrule(r){1-1} \cmidrule(l){2-2}
1     & Conv1d ($k=3$, $o=512$) $\rightarrow$ ReLU \\
2     & Maxpool1d ($k=3$, $s=3$)                   \\
3     & Conv1d ($k=3$, $o=512$) $\rightarrow$ ReLU \\
4     & Maxpool1d ($k=3$, $s=3$)                   \\
5     & Conv1d ($k=3$, $o=512$) $\rightarrow$ ReLU \\
6     & Conv1d ($k=3$, $o=512$) $\rightarrow$ ReLU \\
7     & Linear ($o=$ \#classes)                      \\ \bottomrule
\end{tabular}%
}
\caption{
    Architecture of CLCNN.
    Kernel size $k$, output size $o$, and stride size $s$ was set to the above table.
}
\label{tab:clcnn_arch}
\end{table}

\paragraph{Character-level convolutional neural network (CLCNN).}
Table~\ref{tab:clcnn_arch}　shows the architecture of CLCNN.
We trained CLCNN with the same parameters as in \citet{shimada2016document}.
Similar to training the character embedding model, Adam was used to minimize the cross-entropy error.
We set the learning rate of Adam to 1e-4 and weight decay to 1e-4, train batch size to 256 for the livedoor, and 512 for the newspaper.

In training the CLCNN, we used the GDCE results obtained by the VCE as the input.
For training, $c = 128$ consecutive characters were extracted from the text in the newspaper, and $c = 80$ consecutive characters were extracted from the title text in the livedoor.
For evaluation, in the newspaper, $c = 128$ characters were slid one by one, the entire text was used as input in the same manner as in \citet{shimada2016document}; in the livedoor, it was the same as in the training.
    
\section{Results and Discussion}
    \begin{table*}[t]
\centering
\begin{tabular}{@{}lcrrrrr@{}}
\toprule
                    & \multicolumn{6}{c}{Accuracy [\%]}                                                                                                                                                                                                                  \\ \cmidrule(l){2-7}
                    & \multicolumn{3}{c}{Newspaper}                                                                                    & \multicolumn{3}{c}{Livedoor}                                                           \\ \cmidrule(r){2-4} \cmidrule(l){5-7}
+ CLCNN             & \multicolumn{1}{l}{Vanilla}                           & \multicolumn{1}{l}{+ WT}          & \multicolumn{1}{l}{+ SSA (\textbf{Ours})} & \multicolumn{1}{l}{Vanilla}       & \multicolumn{1}{l}{+ WT}          & \multicolumn{1}{l}{+ SSA (\textbf{Ours})}   \\ \cmidrule(r){1-1} \cmidrule(lr){2-2} \cmidrule(lr){3-3} \cmidrule(lr){4-4} \cmidrule(lr){5-5} \cmidrule(lr){6-6} \cmidrule(l){7-7}
VCE (\textbf{Ours}) & \multicolumn{1}{r}{\textbf{81.02}}                             & 82.78                             & {\footnotesize $^{\dagger}$}\textbf{84.00}        & \textbf{67.16}                             & 68.59                             & {\footnotesize $^{\dagger}$}\textbf{69.05} \\
CAE                 & \multicolumn{1}{r}{{\footnotesize $^{\ddagger}$}79.81} & {\footnotesize $^{\ddagger}$}81.62 & 81.35                                     & {\footnotesize $^{\ddagger}$}58.39 & {\footnotesize $^{\ddagger}$}60.87 & 60.53                              \\ \bottomrule
\end{tabular}
\caption{A comparison between the VCE (with proposed GDCE) and the CAE in the newspaper and the livedoor results. 
We compared our proposed framework (presented as $\dagger$; a disentangled representation) with the state-of-the-art framework of \citet{shimada2016document} (presented as $\ddagger$; without the consideration of disentangled representation).
Our proposed framework had the highest performance.
The model using the VCE performed better than the CAE.}
\label{tab:classification_result}
\end{table*}

First, as a comparison of embedding methods, we compared the GDCE with the conventional CAE-based embedding~\cite{shimada2016document}.
Second, as a comparison of data augmentation methods for image-based character embedding, we also compared the proposed SSA with the conventional WT, the latter of which has reported excellent results but offers no way of interpreting the change on the embedding space.

\subsection{Effectiveness of the Proposal on Text Classification}
Table~\ref{tab:classification_result} presents a comparison of the proposed GDCE and CAE-based embedding.
The GDCE showed better document- and sentence-level classification performance than the conventional CAE-based character embedding without data augmentation.
This may be due to the fact that the characters to be learned by the VCE are distributed in a limited embedded space centered on zeros, so the later stage of the CLCNN training became more effective.
The WT, which randomly set all representations of a particular character embedding to zero, enhanced the discrimination of both models.
The effect on the CAE-based model was particularly large, as reported in previous studies.
We can confirm an effect of the WT as a dropout for preventing overfitting, but it did not provide an interpretation of what was changed in the character embeddings.

The proposed SSA provided us with an idea of what the embedding changes would look like, while also providing the same discriminatory capacity as the WT.
This may be due to the fact that the GDCE had standardized metrics in the embedding space (i.e., the embedding had a normal distribution), so that the distances between the character embeddings were within the range of what could be assumed.
Hence, the size of the perturbations applied could be designed, allowing for meaningful data augmentation.
However, the CAE with SSA did not show an improved classification performance. 
This may be due to the fact that the CAE with SSA does not change to a meaningful character representation.

\begin{figure*}[t]
    \centering
    \begin{minipage}{0.48\linewidth}
        \vspace{2.0mm} 
        \includegraphics[width=\linewidth]{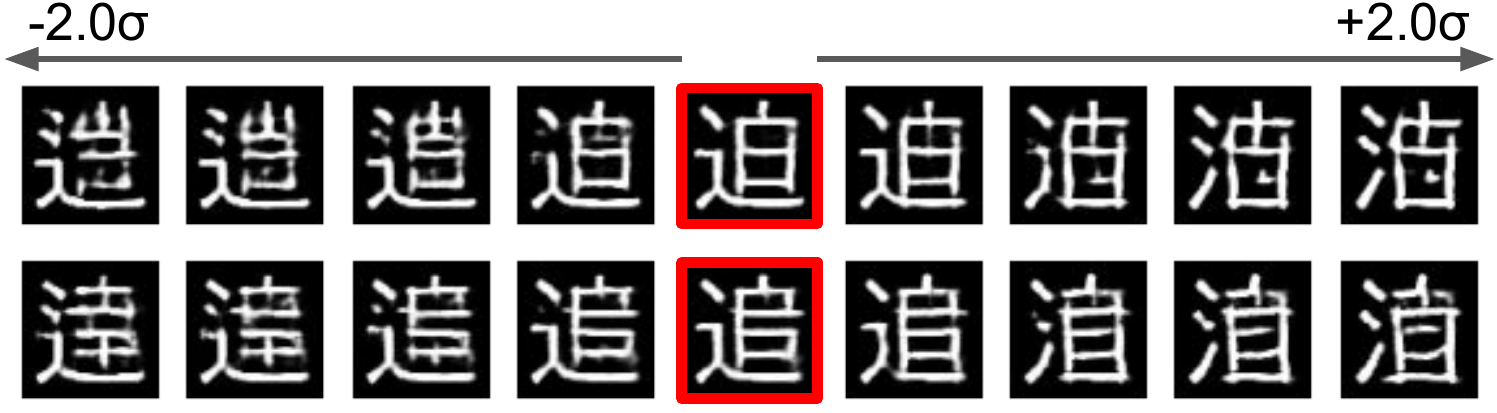}
        \\
        \vspace{-3.0mm} \\
        \includegraphics[width=\linewidth]{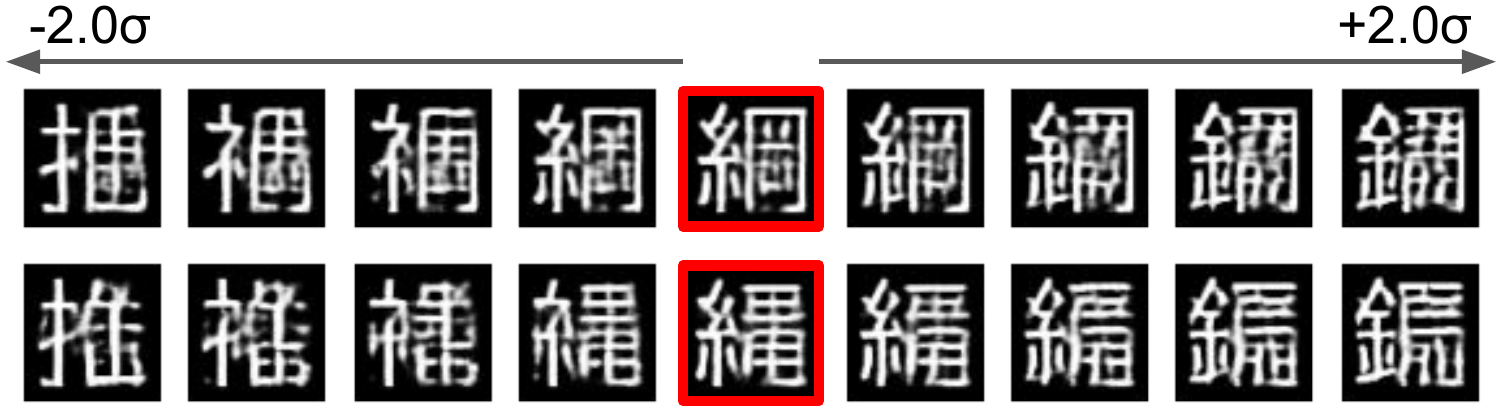}
        \subcaption{VCE (\textbf{proposed})}
        \label{fig:recon_vae}
    \end{minipage}
    \hfill
    \begin{minipage}{0.48\linewidth}
        \vspace{2.0mm} 
        \includegraphics[width=\linewidth]{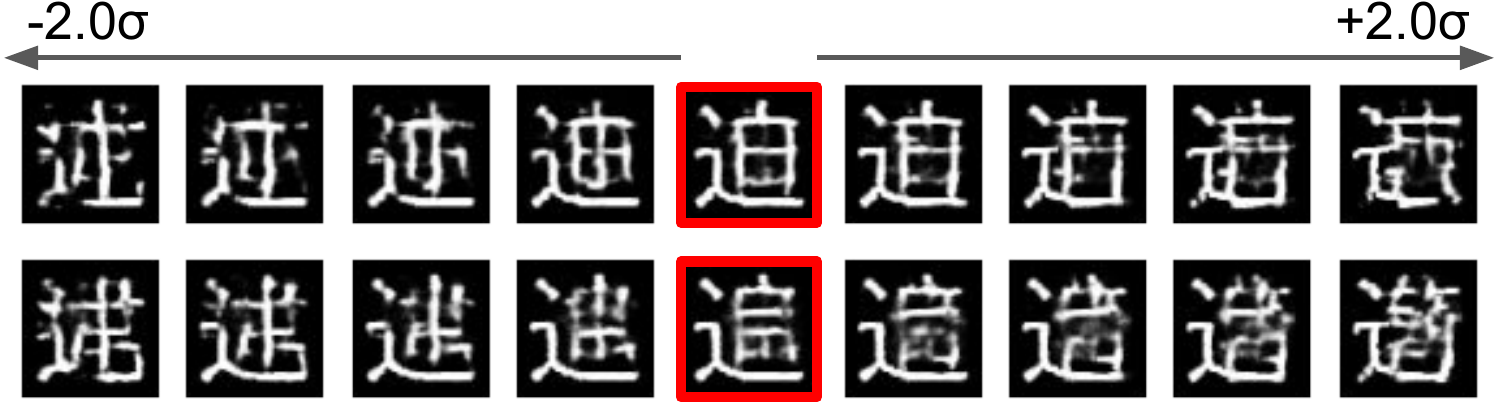}
        \\
        \vspace{-3.0mm} \\
        \includegraphics[width=\linewidth]{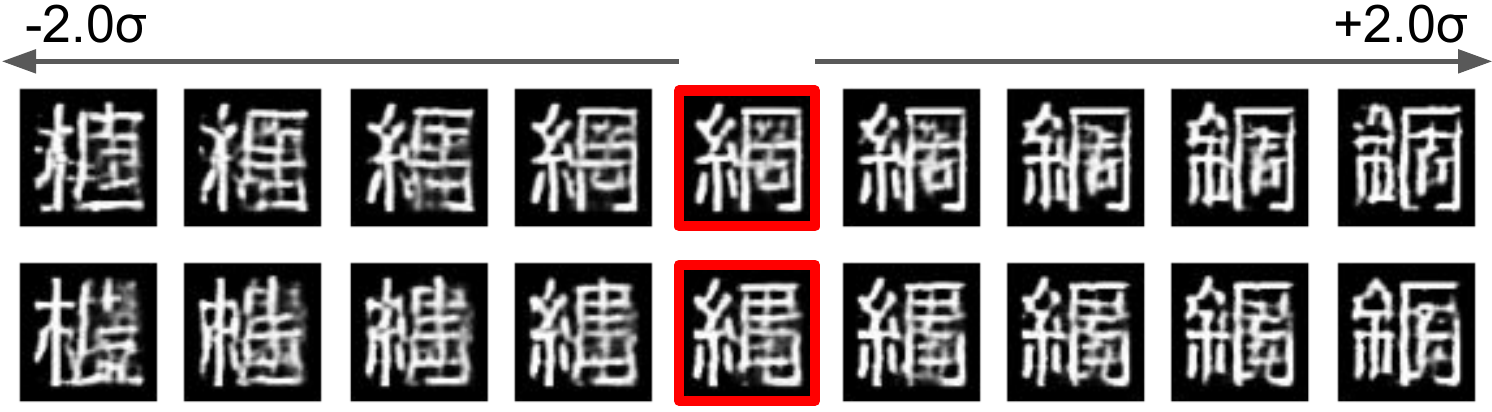}
        \subcaption{CAE}
        \label{fig:recon_cae}
    \end{minipage}

    \caption{
        The results of reconstructing character images from the character embedding trained by the VCE and CAE with perturbation added between $2.0\sigma$.
        \textbf{The upper side} is the reconstructed image of ``迫'' (approach) and ``追'' (follow).
        In the reconstruction from the embedding by the VCE and by adding noise to the fifth dimension of the embedding of ``迫'' or ``追'' (containing a sub-char. of \begin{CJK*}{UTF8}{ipxm}``辶''\end{CJK*} meaning \textit{road}), it can be interpreted that it changed to \begin{CJK*}{UTF8}{ipxm}``氵''\end{CJK*} (sub-char. of \textit{water}) or \begin{CJK*}{UTF8}{ipxm}``辶''\end{CJK*} (sub-char. of \textit{road}, the same as \begin{CJK*}{UTF8}{ipxm}``辶''\end{CJK*}).
        \textbf{The lower side} is the reconstructed image of ``綱'' (rope) and ``縄'' (cord).
        In the reconstruction from the embedding by the VCE and by adding noise to the first dimension of the embedding of ``綱'' or ``縄'' (containing a sub-char. of \begin{CJK*}{UTF8}{ipxm}``糸''\end{CJK*} meaning \textit{yarn}), it can be interpreted that it changed to \begin{CJK*}{UTF8}{ipxm}``扌''\end{CJK*} (sub-char. of \textit{hand}) or \begin{CJK*}{UTF8}{ipxm}``金''\end{CJK*} (sub-char. of \textit{gold}).
    }
    
    \label{fig:recon}
\end{figure*}

\subsection{Effectiveness of the Proposal on Interpretation}
Figure~\ref{fig:recon} shows a comparison of the reconstructed character images when a $\pm2.0\sigma$ perturbation is placed on the \ref{fig:recon_vae} (the GDCE) and \ref{fig:recon_cae} character embedding obtained by the CAE.
In Figure~\ref{fig:recon_vae}, it is confirmed that the shape of the character replaced a different interpretable character or characters with a similar different subcomponent in the input space.
In particular, by adding a perturbation to the fifth dimension of the embedding of ``迫'' or ``追'' (containing a sub-char. of \begin{CJK*}{UTF8}{ipxm}``辶,''\end{CJK*} meaning \textit{road}), it can be interpreted that it changed to \begin{CJK*}{UTF8}{ipxm}``氵''\end{CJK*} (sub-char. of \textit{water}) or \begin{CJK*}{UTF8}{ipxm}``辶''\end{CJK*} (sub-char. of \textit{road}, the same as \begin{CJK*}{UTF8}{ipxm}``辶''\end{CJK*}). 
In addition, by adding a perturbation to the first dimension of the embedding of ``綱'' or ``縄'' (containing a sub-char. of \begin{CJK*}{UTF8}{ipxm}``糸,''\end{CJK*} meaning \textit{yarn}), it can be interpreted that it changed to \begin{CJK*}{UTF8}{ipxm}``扌''\end{CJK*} (sub-char. of \textit{hand}) or \begin{CJK*}{UTF8}{ipxm}``金''\end{CJK*} (sub-char. of \textit{gold}).
From these results, we are convinced that such a replacement in the embedding resulted in more effective data augmentation for training the model.

As seen in Figure~\ref{fig:recon_cae}, in contrast, we were unable to identify these trends.
We consider this is one of the typical benefits of our framework in that each dimension of the GDCE is independent and each of them affects each character component (e.g., sub-char. or radical of the character) with independence.
In other words, we can change only some part of the character by changing certain dimensions of the embedding.

Since the SSA is a local transformation for the parts of the character shown above, even some characters that do not actually exist are generated by the combination of parts.
These are not readable as \textit{correct} characters, but we can make certain interpretations of them.
In sum, the combination of the proposed GDCE and SSA provides us with the interpretability of the data augmentation as well as embedding the character while providing a high discriminative power.

\begin{figure}[t]
    \centering
    \begin{minipage}{\linewidth}
        \centering
        \includegraphics[width=\linewidth]{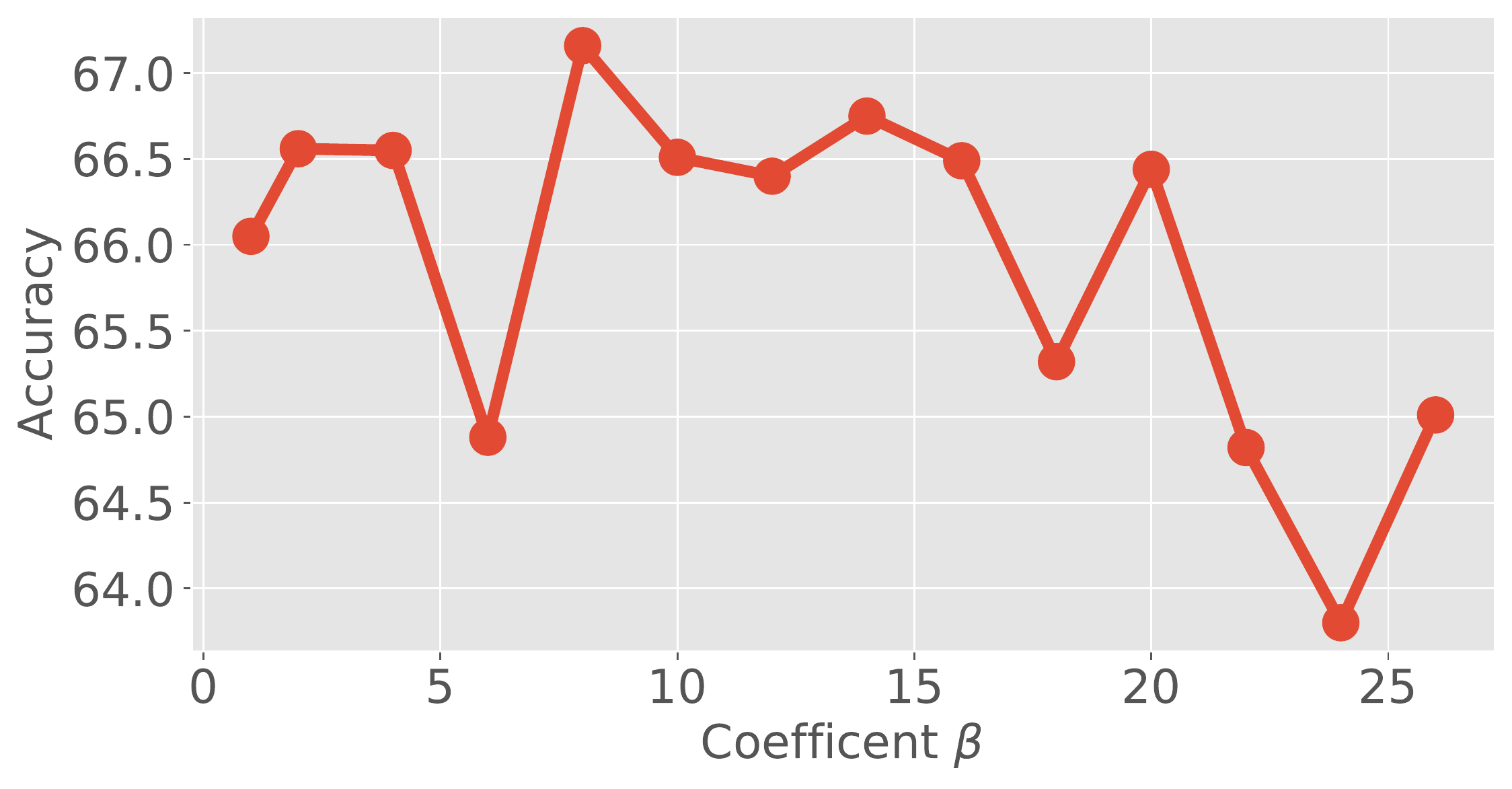}
        \subcaption{
            The effect of coefficient $\beta$ ($\gamma = 0$ i.e., without SSA).
        }
        \label{fig:beta}
    \end{minipage}
    \\
    \vspace{6.5mm}
    \begin{minipage}{\linewidth}
        \centering
        \includegraphics[width=\linewidth]{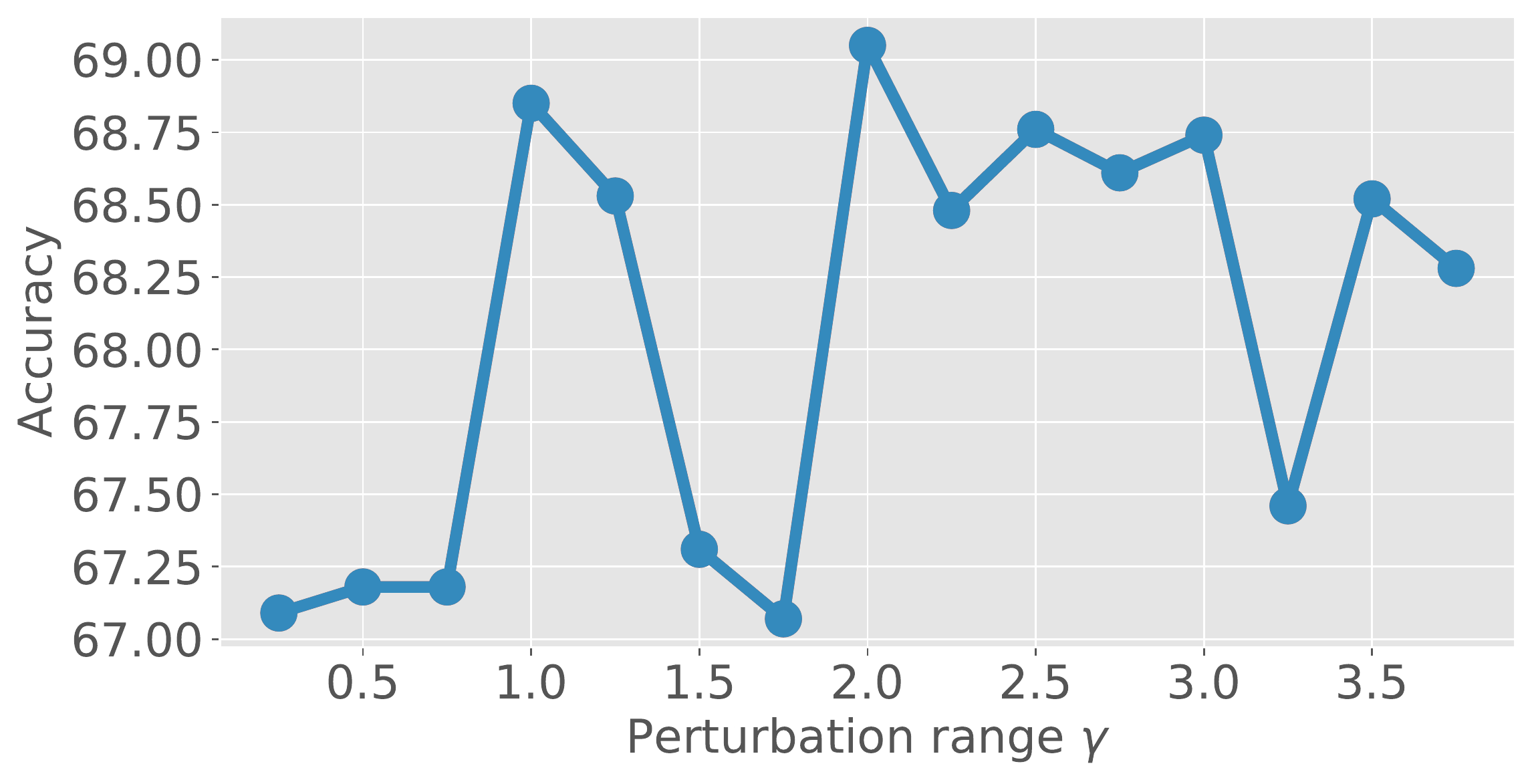}
        \subcaption{
            The effect of perturbation range $\gamma$ ($\beta = 8$).
        }
        \label{fig:gamma}
    \end{minipage}
    \caption{
        The effect of hyperparameters in our framework using the livedoor dataset on the evaluation performance.
    }
    \label{fig:effect_of_hyperparameters}
\end{figure}

\subsection{The Effect of Hyperparameters}\label{sec:appendix_hyparparameters}
To understand the effect of hyperparameters, we analyzed the coefficient $\beta$ and perturbation size $\gamma$ using the livedoor, as shown in Figure~\ref{fig:effect_of_hyperparameters}.

\paragraph{The effect of coefficient $\beta$.}
Figure~\ref{fig:beta} shows the effect of coefficient $\beta$ on the evaluation performance with $\gamma = 0$ (i.e., without SSA).
In our experiments, we confirmed that $\beta = 8$ is the best from the viewpoint of disentanglement and accuracy.

\paragraph{The effect of perturbation size $\gamma$.}
Figure~\ref{fig:gamma} shows the effect of perturbation range $\gamma$ in SSA on the evaluation performance with $\beta = 8$.
Based on the notion that each dimension of the target character embedding follows $\mathcal{N}(\bm{0},\bm{I}_{d'})$, the perturbation range $\gamma$ was chosen to be from $1.0\sigma$ (covering 68\% of the distribution) to $3.0\sigma$ (covering almost the entire distribution).
The best performance was obtained when the perturbation range was set to $\gamma = 2.0$.
This suggests that the character embedding trained by the VCE followed a normal distribution with a mean of $\mu = 0$ and a standard deviation of $\sigma = 1.0$.
To cover the distribution, it is considered useful to add perturbation in the range of $\gamma = 2.0$ corresponding to $2.0\sigma$ (covering 95\% of the distribution).

\subsection{Limitations of the Current Study}
At present, the role of each dimension in the character reconstruction of the GDCE cannot be clearly defined because it depends on the training of the model.
Also, since the VCE was independently trained from the classifier (i.e., not in an end-to-end manner), trained embedding can only consider visual features, not the semantic ones.
We will be working on these in the future.
    
\section{Conclusion}
    We propose a new character-based text classification framework for non-alphabetic languages. 
As the name implies, the combination of our GDCE and SSA not only provided embedding interpretability but also improved the text classification performance.
Our GDCE provided better text classification performance than conventional CAE-based character embedding without data augmentation.
Finally, our framework achieved a competitive result to the conventional state-of-the-art CAE-based embedding with WT while also providing model interpretability.

\bibliography{reference}
\bibliographystyle{acl_natbib}

\end{document}